\documentclass[10pt,twocolumn,letterpaper]{article}

\usepackage[pagenumbers]{iccv} 

\definecolor{iccvblue}{rgb}{0.21,0.49,0.74}
\usepackage[pagebackref,breaklinks,colorlinks,allcolors=iccvblue]{hyperref}

\usepackage{algorithm}
\usepackage{algorithmic}
\usepackage{pifont}
\usepackage{amsmath, amssymb}
\usepackage{multirow}
\usepackage{booktabs}
\usepackage{colortbl}

\newcommand{\qiankun}[1]{\textcolor{black}{#1}}

\def\paperID{1724} 
\def\confName{ICCV}
\def\confYear{2025}

\title{MonoMVSNet: Monocular Priors Guided Multi-View Stereo Network}

\author{Jianfei Jiang, Qiankun Liu\thanks{Corresponding author}, Haochen Yu, Hongyuan Liu, Liyong Wang, Jiansheng Chen, Huimin Ma\textsuperscript{*}\\
University of Science and Technology Beijing, China\\
{\tt\small {\{jiangjf,haochen.yu,hongyuanliu,wangly\}}@xs.ustb.edu.cn, {\{liuqk3,jschen,mhmpub\}}@ustb.edu.cn}
}

\begin{document}
\maketitle
\begin{abstract}
Learning-based Multi-View Stereo (MVS) methods aim to predict depth maps for a sequence of calibrated images to recover dense point clouds. However, existing MVS methods often struggle with challenging regions, such as textureless regions and reflective surfaces, where feature matching fails. In contrast, monocular depth estimation inherently does not require feature matching, allowing it to achieve robust relative depth estimation in these regions. To bridge this gap, we propose MonoMVSNet, a novel monocular feature and depth guided MVS network that integrates powerful priors from a monocular foundation model into multi-view geometry. Firstly, the monocular feature of the reference view is integrated into source view features by the attention mechanism with a newly designed cross-view position encoding. Then, the monocular depth of the reference view is aligned to dynamically update the depth candidates for edge regions during the sampling procedure. Finally, a relative consistency loss is further designed based on the monocular depth to supervise the depth prediction. Extensive experiments demonstrate that MonoMVSNet achieves state-of-the-art performance on the DTU and Tanks-and-Temples datasets, ranking first on the Tanks-and-Temples Intermediate and Advanced benchmarks. 
The source code is available at \href{https://github.com/JianfeiJ/MonoMVSNet}{https://github.com/JianfeiJ/MonoMVSNet}.
\end{abstract}    
\section{Introduction}

\begin{figure}[!t]
  \centering
  \begin{minipage}{\linewidth}
    \centering
    \includegraphics[width=\linewidth]{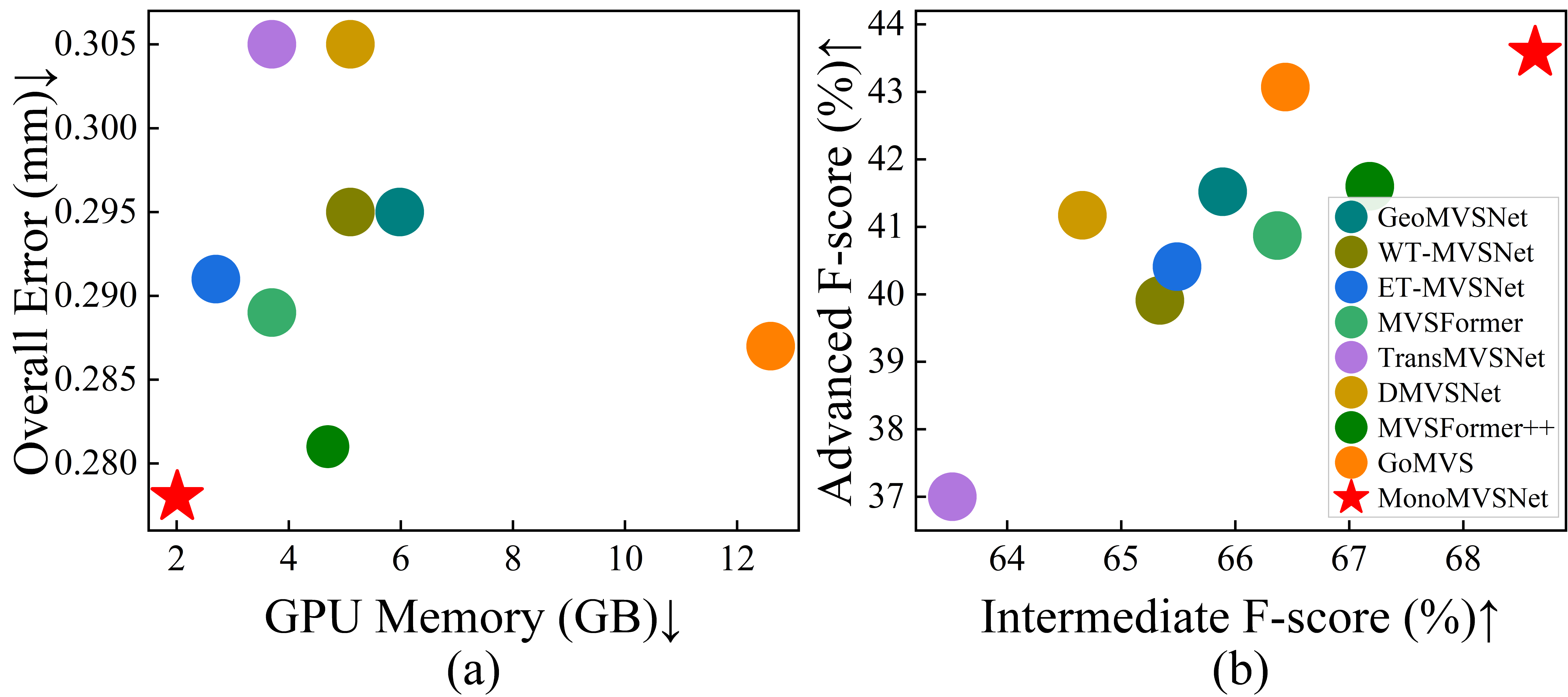}
  \end{minipage}

  \begin{minipage}{\linewidth}
    \centering
    \includegraphics[width=\linewidth]{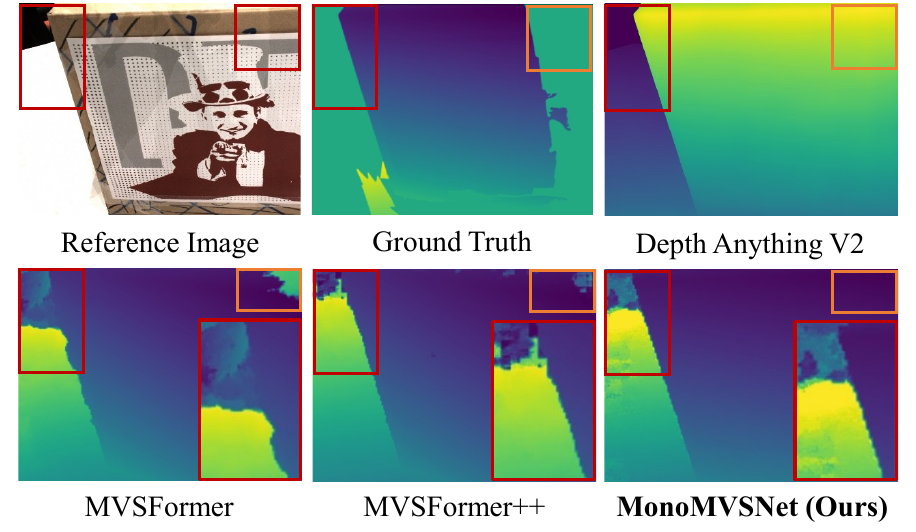}
  \end{minipage}

  \caption{\textbf{Row 1}: (a) Comparison with SOTA methods in terms of overall error and GPU memory consumption at a resolution of 832$\times$1152 with 5-view images on the DTU \cite{dtu} test set, where \textbf{lower is better}; (b) Comparison with SOTA methods on the Tanks-and-Temples \cite{tanks} benchmark, where \textbf{higher is better}. \textbf{Row 2-3}: Qualitative depth comparison with Depth Anything V2 \cite{dav2}, MVSFormer \cite{mvsformer}, and MVSFormer++ \cite{mvsformer++} on scan13 from DTU test set. Our method produces more accurate metric depth in edge (red bounding box) and textureless (orange bounding box) regions.}
  \label{fig:compare}
  \vspace{-4mm}
\end{figure}

Multi-View Stereo (MVS) is a fundamental task in computer vision that aims to reconstruct dense 3D geometry~\cite{liu2025protocar, yu2024get3dgs, zhao2025sam2object} from a set of calibrated images. While traditional MVS methods \cite{gipuma, acmp, acmmp, yuan2025sed, yuan2025dvp, yuan2025msp} have made significant progress, they remain limited by handcrafted feature representations. The rapid advancement of deep learning in recent years has driven the development of learning-based MVS \cite{mvsnet, rmvsnet, casmvsnet, dimvs, jiang2025rrt}, which leverage the powerful representation capabilities of deep neural networks, leading to substantial performance gains over traditional approaches.

MVS can essentially be understood as a one-to-many feature matching problem \cite{transmvsnet}. However, feature matching is often challenged by difficult regions, such as texture-less areas, reflective surfaces, and depth discontinuity edges, leading to suboptimal reconstruction results, as shown in Fig.~\ref{fig:compare}. To mitigate this issue, recent MVS methods have focused on enhancing feature extraction, as improved features enable more effective feature matching. The conventional approach for feature extraction involves using a Feature Pyramid Network (FPN) \cite{fpn} to extract multi-scale features, facilitating the construction of multi-level cost volumes for coarse-to-fine depth prediction. Building on this, recent methods have integrated deformable convolutions \cite{dcn, transmvsnet, gbinet}, attention mechanisms \cite{transmvsnet, wtmvsnet, etmvsnet}, and pre-trained Vision Transformers (ViTs) \cite{transformer, mvsformer, mvsformer++} to extract more reliable feature representations.

On the other hand, monocular depth estimation has seen significant progress in recent years \cite{dav1, dav2}. Unlike multi-view depth estimation, monocular depth models leverage neural networks to learn contextual cues for depth prediction, inherently mitigating mismatches in challenging regions. Recent monocular foundation models trained on large-scale real-world datasets exhibit strong zero-shot generalization capabilities. However, pre-trained monocular foundation models typically provide only relative depth and lack the geometric constraints of multi-view stereo, limiting their utility in downstream tasks. As illustrated in Fig. \ref{fig:compare} (row 2), while the monocular depth estimation model (\text{i.e.}, Depth Anything V2 \cite{dav2}) generates visually compelling results, a significant scale ambiguity persists between the predicted depth and the ground truth.

To address this, we propose MonoMVSNet, which integrates the powerful priors from the monocular foundation model with multi-view geometry to construct a more robust and stronger
MVS network. Although previous works \cite{mvsformer,mvsformer++} have explored the usage of pre-trained ViTs to improve feature representations in multi-view stereo, they
rely on complex training strategies and architectures that require pre-trained ViT features for all input views, introducing significant overhead. In contrast, our approach uses a simpler architecture, achieving better performance with lower GPU memory consumption, as shown in Fig. \ref{fig:compare}.

To fully exploit the strong generalization ability of the monocular foundation model, we first utilize pre-trained monocular features for robust feature extraction. Specifically, we extract monocular features only from the reference view and combine them with FPN features. 
Subsequently, for the reason that traditional positional encoding is designed for 2D images and does not capture 3D spatial relationships across viewpoints, we propose a novel Cross-View Position Encoding (CVPE) tailored for MVS to enhance both intra-view and inter-view attention mechanisms, effectively improving the representational capacity of source features. Furthermore, monocular depth provides fine-grained 
relative depth information in edge regions, which is essential for capturing depth discontinuities. To maximize the utility of this information, we introduce a monocular depth alignment module that aligns monocular relative depth with the predicted depth and guides the dynamic depth sampling process, ensuring a more accurate selection of depth candidates. Finally, we propose a relative consistency loss to enforce consistency between the aligned monocular depth and the predicted depth.

In summary, our contributions are as follows:
\begin{itemize}
\item We design a simple and effective approach that utilizes monocular feature priors to construct a powerful feature extractor. The proposed cross-view position encoding significantly improves the efficiency of feature exchange between different views.
\item We design a dynamic depth sampling strategy using monocular relative depth priors and enforce relative depth consistency through a relative consistency loss, thereby improving the representation of depth discontinuities.
\item MonoMVSNet achieves state-of-the-art performance on the DTU dataset and the Tanks-and-Temples benchmark.
\end{itemize}

\label{sec:intro}

\section{Related Work}

\begin{figure*}[!t]
  \centering
   \includegraphics[width=1.0\textwidth]{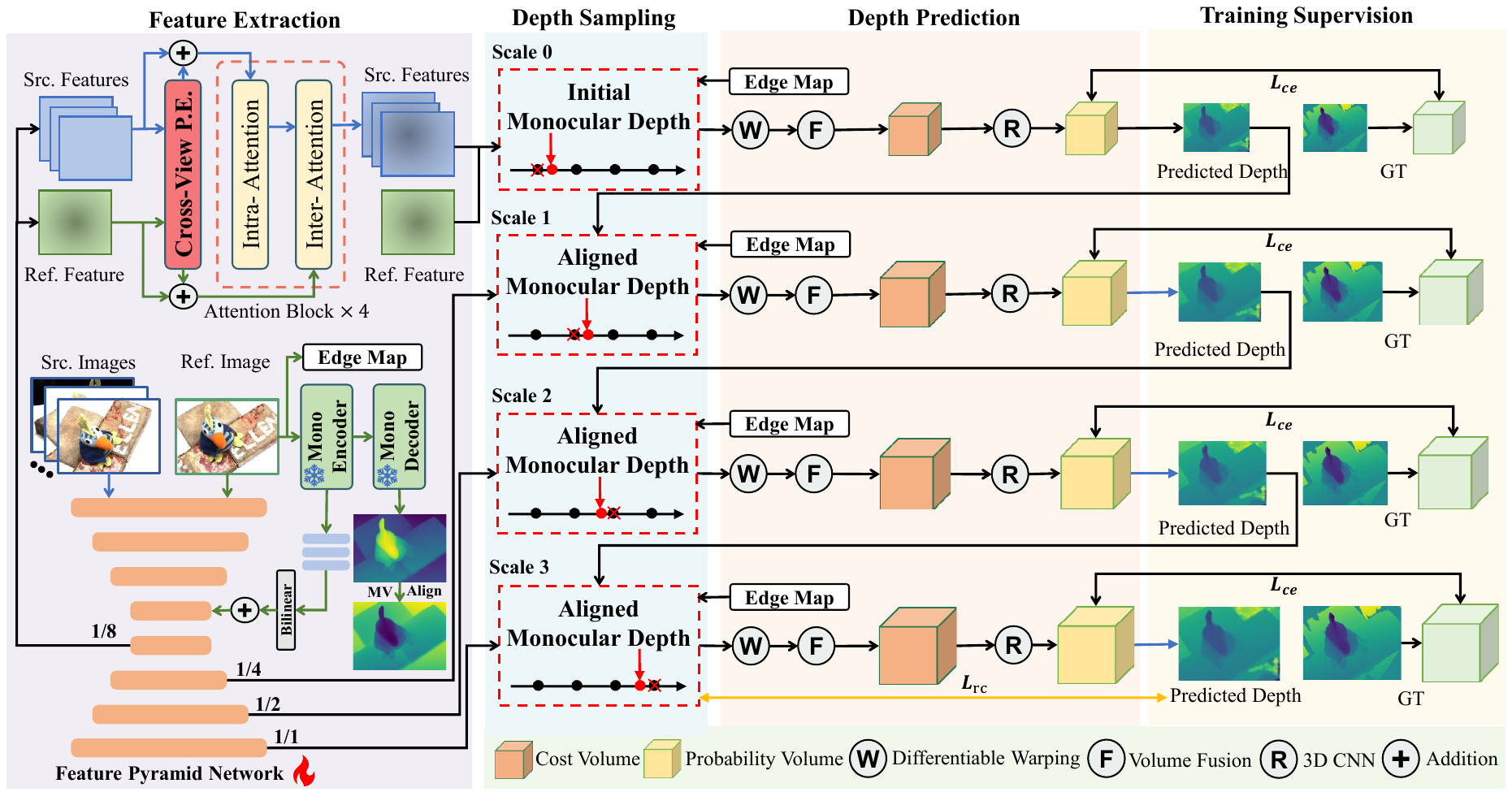}

   \caption{\textbf{Overview of the proposed MonoMVSNet.} \qiankun{(1) Exploitation of monocular feature: The reference monocular feature, extracted by the mono encoder model, is used to enhance the reference FPN feature and integrated into source features by attention mechanism with cross-view position encoding. (2) Exploitation of monocular depth: The monocular depth of the reference image, output by the mono decoder, is aligned to guide the depth sampling procedure and supervise the predicted depth with relative consistency loss.}}
   \label{fig:overview}
\end{figure*}

\noindent{\textbf{Learning-based Multi-View Stereo (MVS).}} MVS aims to reconstruct dense 3D representations from multiple images captured from different viewpoints. MVSNet \cite{mvsnet}, the first end-to-end learning-based MVS method, consists of four main steps: feature extraction, cost volume construction, cost volume regularization, and depth prediction. However, the high memory demand of 3D CNNs for cost volume regularization remains a limitation. Subsequent methods \cite{rmvsnet, d2hcrmvsnet, aarmvsnet} reduced memory demands with RNN-based regularization, while cascade-based approaches \cite{casmvsnet, ucsnet, cvpmvsnet} improved efficiency via coarse-to-fine processing. However, existing methods still struggle with reconstruction quality, emphasizing the need for better feature representations and depth sampling strategies.

\noindent{\textbf{Feature Representation in MVS.}} Effective feature representation is crucial for MVS performance. Cascade-based MVS methods \cite{casmvsnet} leverage FPN \cite{fpn} for multi-scale feature extraction. To further improve this, some studies \cite{transmvsnet, wtmvsnet, etmvsnet} incorporates Transformers \cite{transformer} to aggregate global features through intra- and inter-view attentions. MVSFormer \cite{mvsformer} fine-tunes a pre-trained Vision Transformer (ViT) \cite{twins} on high-resolution images for better feature representations. Building on this, MVSFormer++ \cite{mvsformer++} further improves feature learning by injecting cross-view information into the pre-trained DINOv2 \cite{dinov2}. In contrast to these methods, we integrate monocular features from pre-trained monocular foundation models \cite{dav1, dav2} with FPN features for the reference view only, and transfer these features to source views using cross-view position encoding to efficiently enhance feature representation robustness, while greatly reducing overhead.

\noindent{\textbf{Depth Sampling in MVS.}} In addition to feature extraction, depth sampling strategies are crucial for constructing accurate cost volumes in MVS methods. These methods construct the cost volume by sampling multiple depth candidates across the depth range and warping source feature maps onto corresponding depth hypothesis planes in the reference view. MVSNet \cite{mvsnet} performs dense depth sampling, which is inefficient. CasMVSNet \cite{casmvsnet} introduces a coarse-to-fine depth sampling strategy. Building on this strategy, later studies have to further enhance memory efficiency and inference speed. For instance, 
IS-MVSNet \cite{ismvsnet} incorporates an importance sampling module, 
GBi-Net \cite{gbinet} employs a generalized binary search strategy, and MaGNet \cite{gbinet} samples based on the monocular depth probability distribution. These methods focus on sampling fewer depth candidates to improve memory efficiency and inference speed. In contrast, we utilize monocular depth to dynamically guide the depth sampling process, obtaining more reasonable depth candidates for better depth estimation, and thereby enhancing reconstruction performance.

\noindent{\textbf{Monocular Depth Estimation (MDE).}} Despite improvements in MVS methods, accurately reconstructing regions such as textureless areas, depth discontinuities, and reflective surfaces remains challenging. Recently, monocular depth estimation has made substantial progress. MiDaS \cite{midas} pioneered zero-shot generalization, while DPT \cite{dpt} leveraged Transformers for fine-grained estimation. Depth Anything V1 \cite{dav1} leverages a large number of unlabeled images, overcoming the traditional issue of insufficient labeled data. Depth Anything V2 \cite{dav2} further incorporates knowledge distillation, allowing for robust depth prediction across complex scenarios. However, the depth maps generated by these monocular depth estimation methods often suffer from scale ambiguity, making them unsuitable for direct use in downstream tasks (\textit{e.g.}, 3D reconstruction). To address this, we combine the strengths of monocular depth estimation and multi-view stereo to achieve robust multi-view depth estimation in challenging regions.

\section{Methodology}

\subsection{Overview}
The framework of MonoMVSNet is depicted in Fig. \ref{fig:overview}, which efficiently integrates monocular features and depth from a pre-trained monocular foundation model \cite{dav2}. For the usage of monocular feature, we only feed the monocular model with the reference image to get the reference monocular feature, which avoids the overhead introduced by the monocular model as much as possible and also boosts the model performance. The reference monocular feature is integrated into source features through the attention mechanism, which is enhanced by the newly designed cross-view position encoding (Section~\ref{sec:feature_extract}). For the usage of monocular depth, we first align it with the coarse depth produced by MonoMVSNet by filtering unreliable positions. The aligned monocular depth is used to replace the depth candidates around the edge regions in the depth sampling procedure (Section~\ref{sec:mono_depth}) and to supervise the depth produced by MonoMVSNet with relative consistency loss (Section~\ref{sec:training_objective}). 

\subsection{Monocular Feature for Feature Extraction}
\label{sec:feature_extract}
Given $N$ input images $ \left\{\mathbf{I}_n\right\}_{n=0}^{N-1} \in \mathbb{R}^{3 \times H \times W} $, consisting of a reference image (denoted as $\mathbf{I}_0$) and $N-1$ source images, our goal is to estimate a depth map for the reference image using these input images and their corresponding camera parameters. Here, 
$H$ and $W$ represent the height and width of the input images, respectively.

\noindent{\textbf{Feature Extraction for Different Images.}}
We employ a 4-layer FPN to extract multi-scale FPN features for each of the source images. Let $s$ be the scale index and $C$ be the channel dimensionality of FPN, the FPN feature of the $N-1$ source images can be denoted as $\{\mathbf{F}_{n,s} \in \mathbb{R}^{C\times \frac{H}{2^{3-s}} \times \frac{W}{2^{3-s}}}|s=0,1,2,3\}_{n=1}^{N-1}$. 

To extract features for the reference image, we first feed it to the pre-trained monocular model (specifically, Depth Anything V2~\cite{dav2}), which produces the reference monocular feature $\mathbf{F}_0^{mono} \in \mathbb{R}^{C' \times h \times w}$, where $C'$ is the channel dimensionality of the pre-trained monocular model and $(h, w)$ is the spatial resolution of the feature. At the same time, the reference image is also fed into the encoder of FPN, producing the feature map of the lowest resolution, which is denoted as $\mathbf{F}_{0}^{enc} \in \mathbb{R}^{C\times \frac{H}{8} \times \frac{W}{8}}$. The feature $\mathbf{F}_0^{mono}$ is processed by a convolution layer and a bilinear upsampling operation to align the channel dimensionality and spatial resolution with $\mathbf{F}_{0}^{enc}$, which are added with each other:
\begin{equation}
    \mathbf{F}_{0} = \mathbf{F}_{0}^{enc} \oplus {\rm Bilinear}({\rm Conv}(\mathbf{F}_{0}^{mono})).
\end{equation}
The feature $\mathbf{F}_{0}$ is further fed into the FPN decoder to get the multi-scale features for the reference image $\mathbf{I}_0$, which is denoted as $\{\hat{\mathbf{F}}_{0,s} \in \mathbb{R}^{C\times \frac{H}{2^{3-s}} \times \frac{W}{2^{3-s}}}|s=0,1,2,3\}$.

In our design, the overhead introduced by the pre-trained monocular model is reduced as much as possible since only the reference image is processed by the monocular model. However, only the reference feature contains priors from the monocular model. To solve this, we use the attention mechanism to enhance source features with a newly designed cross-view position encoding, which also integrates the monocular prior from the reference feature to source features. As we will see in Section~\ref{sec:ablation_study}, such design can also boost the overall performance of MonoMVSNet compared to the peer model design that extracts and integrates the monocular feature for all source images.

\noindent\textbf{Cross-View Position Encoding for Attention.}
Exploiting the intra-view and inter-view attention to enhance source features is a common practice in existing MVS works~\cite{transmvsnet, wtmvsnet, etmvsnet}. 
Though the source features can be enhanced to some extent in these works, they usually use the relative or absolute position encoding for attention, neglecting the importance of 3D spatial information across different views. To alleviate this, we propose a novel Cross-View Position Encoding (CVPE) to strengthen intra-view and inter-view attention interactions, which integrates the priors from the reference feature to source features more effectively.

\begin{figure}[t]
  \centering
   \includegraphics[width=1.0\linewidth]{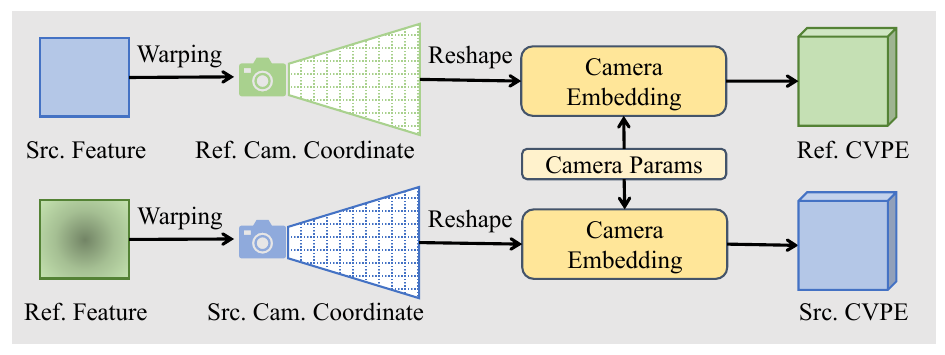}

   \caption{\textbf{Illustration of the Cross-View Position Encoding (CVPE).} Each pair of reference and source features is warped into the respective views. Together with the camera parameters, they undergo camera embedding to generate the CVPE for both the reference and source features.}
   \label{fig:cvpe}
   \vspace{-2mm}
\end{figure}

Supposing the depth hypotheses in the 0-th scale are $\{d_{i,0}\}_{i=0}^{D_0 - 1}$, the $n$-th source feature $\mathbf{F}_{n,0}$ is warped to these $D_0$ hypothesis planes using camera intrinsic matrix $\mathbf{K_n}$ of the $n$-th source view, rotation matrix $\mathbf{R}_{n \rightarrow 0}$ and translation vector $\mathbf{t}_{n \rightarrow 0}$ from the $n$-th source view to the reference view. We denote the warped 
feature as $\mathbf{F}_{n\rightarrow 0,0} \in \mathbb{R}^{C \times D_0 \times \frac{H}{8} \times \frac{W}{8}}$. Given the 2D coordinate $\mathbf{c}_{n,0}$ in the $n$-th source feature $\mathbf{F}_{n,0}$, the warped 2D coordinate $\mathbf{c}^{i}_{n\rightarrow0,0}$ on the $i$-th depth hypothesis plane is determined by:
\begin{equation}
\label{eq:project_coord}
    \mathbf{c}^{i}_{n\rightarrow0,0} = \mathbf{K}_0  (\mathbf{R}_{n \rightarrow 0} \mathbf{K}_{n}^{-1} \mathbf{c}_{n,0}  d_{i,0} + \mathbf{t}_{n \rightarrow 0}),
\end{equation}
which means that:
\begin{equation}
\label{eq:replace_feature}
    \mathbf{F}_{n\rightarrow 0,0}[:,i,\mathbf{p}^{i}_{n\rightarrow0,0}] = \mathbf{F}_{n,0}[:,\mathbf{p}_{n,0}],
\end{equation}
where $[\cdot,...,\cdot]$ is the indexing operation of features. Similarly, the reference feature $\hat{\mathbf{F}}_{0,0}$ is also warped to the $n$-th source view, producing the feature  $\hat{\mathbf{F}}_{0\rightarrow n,0} \in \mathbb{R}^{C \times D_0 \times \frac{H}{8} \times \frac{W}{8}}$.
During the feature projection procedure, the frustum-shaped features may lead to information loss in the spatial dimension. To mitigate this, we employ a camera embedding module composed of a multi-layer perceptron and a squeeze-and-excitation layer~\cite{selayer} to encode the imag-to-world camera parameters of both the reference and source views into the projected features. This process effectively injects camera parameter-related spatial information into the features and compresses the features along the depth dimension, which produces the CVPE $\mathbf{F}'_{n\rightarrow 0,0} \in \mathbb{R}^{C \times \frac{H}{8} \times \frac{W}{8}}$ and $\hat{\mathbf{F}}'_{0\rightarrow n,0} \in \mathbb{R}^{C \times \frac{H}{8} \times \frac{W}{8}}$, as shown in Fig.~\ref{fig:cvpe}.

The reference CVPE $\mathbf{F}'_{n\rightarrow 0,0}$ and source CVPE $\mathbf{F}'_{0\rightarrow n,0}$ are added to their corresponding features, which are fed into the intra-view and inter-view attention blocks. The output of the final attention block is the enhanced feature $\hat{\mathbf{F}}_{n,0}$ for the $n$-th view in the 0-th scale. Finally, the enhanced feature is further processed by a convolution layer and a bilinear upsampling operation to be added to the next scale of the FPN feature, producing the enhanced feature for the next scale. We denote the enhanced features of all source views as $\{\hat{\mathbf{F}}_{n,s} \in \mathbb{R}^{C\times \frac{H}{2^{3-s}} \times \frac{W}{2^{3-s}}}|s=0,1,2,3\}_{n=1}^{N-1}$.

\subsection{Monocular Depth for Dynamic Depth Sampling}
\label{sec:mono_depth}


The monocular depth $\mathbf{D}^{mono} \in \mathbb{R}^{H\times W}$ produced by the monocular model suffers from scale ambiguity, making it unsuitable for direct application in MVS tasks. To address this, an alignment method is designed. 
Since the monocular model is more robust in challenging regions, we exploit the monocular depth to guide the depth sampling procedure with dynamic adjustment for edge regions.

\noindent\textbf{Monocular Depth Alignment.}
Let $\mathbf{D}_s \in \mathbb{R}^{\frac{H}{2^{3-s}} \times \frac{W}{2^{3-s}}}$ be the predicted depth by MonoMVSNet in the $s$-th scale (refer to Section~\ref{sec:cost_volume}), we use $\mathbf{D}_{s-1}$ to align the monocular depth $\mathbf{D}^{mono}$ for the $s$-th stage. Note that in the 0-th scale, the monocular depth is directly scaled to fit the minimum and maximum depth range based on the predefined depth range.

Take the $s$-th ($s\geq 1$) scale for example, we first bilinearly resize the monocular depth $\mathbf{D}^{mono}$ and the depth predicted in the previous scale $\mathbf{D}_{s-1}$ to the resolution $(\frac{H}{2^{3-s}}, \frac{W}{2^{3-s}})$, which are respectively denoted as $\hat{\mathbf{D}}^{mono}_{s}$ and $\hat{\mathbf{D}}_{s}$. Since the depth predicted in the previous scale is coarser than that in the current scale, it is necessary to filter out the depth of unreliable pixels in $\hat{\mathbf{D}}_{s}$ before aligning $\hat{\mathbf{D}}^{mono}_{s}$ to $\hat{\mathbf{D}}_{s}$. To do this, we keep the coordinates of the top 80\% pixels that have the largest confidence scores according to the confidence map produced in the previous scale, and the coordinates set is denoted as $\mathcal{C}_s$. The scale $a$ and shift $b$ for the alignment between $\hat{\mathbf{D}}^{mono}_{s}$ and $\hat{\mathbf{D}}_{s}$ are estimated using the least squares optimization based on these kept pixels:
\begin{equation}
(a, b)=\arg \min _{a, b} \sum_{\mathbf{c} \in \mathcal{C}_s} \left( \frac{1}{\hat{\mathbf{D}}_{s}[\mathbf{c}]}-\left(a \hat{\mathbf{D}}^{mono}_{s}[\mathbf{c}]+b\right)^2\right).
\end{equation}
The aligned monocular depth $\mathbf{D}$ is obtained by:
\begin{equation}
    \mathbf{D}_{s}^{align} = a\hat{\mathbf{D}}^{mono}_{s} + b.
\end{equation}
In the following, we show how to use the aligned depth to guide the depth sampling.

\noindent{\textbf{Monocular Dynamic Depth Sampling.}}
Following previous methods \cite{mvster}, we adopt an inverse depth sampling strategy to sampled depth candidates within the inverse depth range, satisfying the equation $\frac{1}{R_s}=\frac{1}{D_{s-1}-1} \frac{1}{R_{s-1}}$, where $R_s$ and $D_s$ denote the depth range and the number of depth candidates at the $s$-th scale, respectively.

The traditional inverse depth sampling method does not capture relative depth information, leading to blurry depth estimations in discontinuous regions, such as edges. In contrast, monocular models demonstrate impressive zero-shot performance on relative depth estimation. Intuitively, we utilize monocular depth at edge regions to guide depth sampling, ensuring more reasonable depth candidates. 
Specifically, we first apply a lightweight edge estimation network \cite{edge} to predict the edge confidence map $\mathbf{E} \in \mathbb{R}^{H\times W}$ for the reference image, which is bilinearly resized to $\hat{\mathbf{E}}_s \in \mathbb{R}^{\frac{H}{2^{s-1}} \times{\frac{W}{2^{s-1}}}}$ for the $s$-th scale. Only the pixels that have the edge confidence larger than a pre-defined threshold $\lambda$ are kept, and their coordinates set is denoted as $\mathcal{C}_s^{edge}$. 

Supposing the depth hypotheses for the $s$-th scale are $\{d_{i,s}\}_{i=0}^{D_s - 1}$ during the inverse depth sampling procedure. For each pixel with the coordinate $\mathbf{c} \in \mathcal{C}_s^{edge}$, the absolute difference between the depth value $\mathbf{D}_{s}^{align}[\mathbf{c}]$ to all $D_s$ depth candidate values are computed. The depth candidate value that has the smallest absolute difference is replaced by the value $\mathbf{D}_{s}^{align}[\mathbf{c}]$. The remaining pixels whose coordinates are not in $\mathcal{C}_s^{edge}$ share the same depth candidates $\{d_{i,s}\}_{i=0}^{D_s - 1}$. As we can see, the depth candidates are dynamically updated for the edge regions, which captures accurate relative depth information from the monocular depth.

\begin{figure*}[!t]
  \centering
   \includegraphics[width=1.0\textwidth]{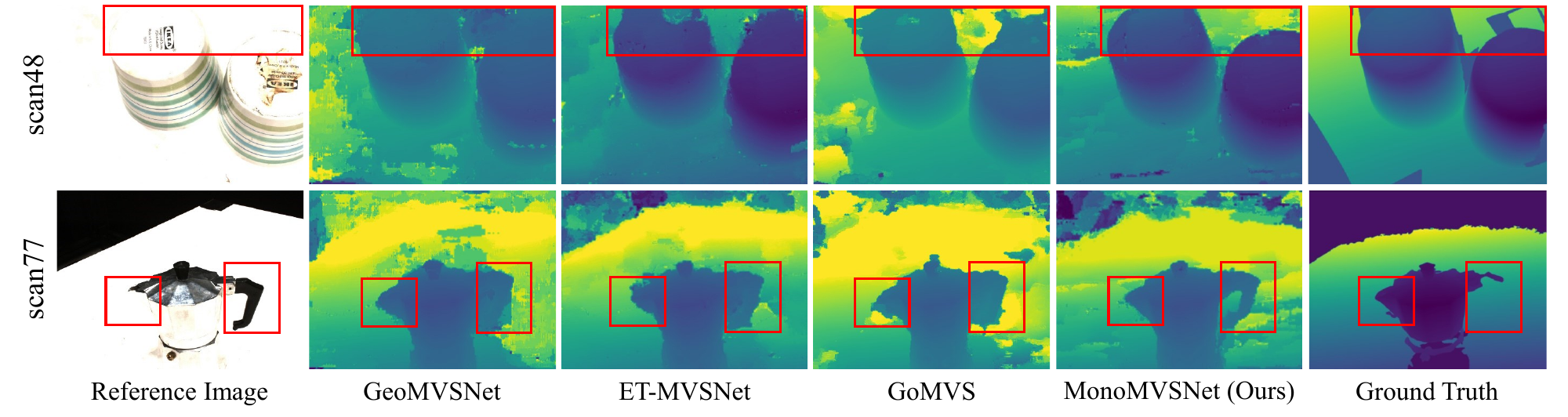}

   \caption{Qualitative comparison of predicted depth maps with GeoMVSNet \cite{geomvsnet}, ET-MVSNet \cite{etmvsnet} and GoMVSNet \cite{gomvs} on the most challenge scenes, scan48 and scan77 from DTU test set. Our method predicts more accurate depth in reflective surfaces and depth-discontinuous edge regions.}
   \label{fig:depth_compare_big}
   \vspace{-2mm}
\end{figure*}

\subsection{Depth Prediction with Cost Volume}
\label{sec:cost_volume}

Following existing works~\cite{gwcnet, cider},
the depth $\mathbf{D}_s$ in the $s$-th scale is obtained with a cost volume, which is constructed based on the correlations between the reference feature and all the warped source features. The overall procedure can be divided into 3 steps: (1) warping all source features from the source view to the reference view, (2) constructing cost volume with feature correlations, and (3) predicting the depth based on the cost volume.
For more details about the construction of cost volume, please refer to the works in \cite{mvster, geomvsnet}. Let $\mathbf{V}_s \in \mathbb{R}^{G \times D_s \times \frac{H}{2^{3-s}} \times \frac{W}{2^{3-s}}}$ be the constructed cost value, it is fed into a lightweight 3D UNet regularization network \cite{mvster} (followed by a softmax function) to obtain the probability volume $\mathbf{P}_s \in \mathbb{R}^{D_s \times \frac{H}{2^{3-s}} \times \frac{W}{2^{3-s}} }$. The probability volume $\mathbf{P}_s$ is then used to predict the depth $\mathbf{D}_s$ through a winners-take-all strategy. Take the pixel with coordinate $\mathbf{c}$ for example:
\begin{equation}
\mathbf{D}_s[\mathbf{c}] = d_{i,s} \quad {\rm s.t.} \quad  \arg \max _{i} \mathbf{P}[i,\mathbf{c}].
\end{equation}

\subsection{Training Objective with Relative Consistency}
\label{sec:training_objective}


To further leverage the outstanding relative depth estimation capability of the monocular model, we propose a relative consistency loss to supervise the the depth produced by MonoMVSNet. 
Take the $s$-th scale for example, we first get the depth with probability $\mathbf{D}_{s}^{prob} \in \mathbb{R}^{\frac{H}{2^{3-s}} \times \frac{W}{2^{3-s}}}$. Take the pixel with coordinate $\mathbf{c}$ for example:
\begin{equation}
\mathbf{D}_s^{prob}[\mathbf{c}] =  \sum_{i=0}^{D_s} \mathbf{P}[i,\mathbf{c}] d_{i,s}.
\end{equation}
Then we randomly sample two sets of pixels with each set contains $M$ pixels. The corresponding coordinate sets are denoted as $\mathcal{C}_s^{1}$ and $\mathcal{C}_s^{2}$. The relative consistency loss is:
\begin{equation}
    \mathcal{L}^{rc}_s = \frac{1}{M} \sum _{m=0}^{M-1} \max (0, - e_m),
\end{equation}
where $e_m$ is the relative error between the $m$-th pixel in these two sets:

\begin{equation}
\begin{split}
 e_m = & (\mathbf{D}_s^{prob}[\mathbf{c}^1_m] - \mathbf{D}_s^{prob}[\mathbf{c}^2_{m}] \cdot \\ &{\rm Sign}(\hat{\mathbf{D}}^{mono}_{s}[\mathbf{c}^1_m] - \hat{\mathbf{D}}^{mono}_{s}[\mathbf{c}^2_m]),
 \end{split}
\end{equation}
where $\mathbf{c}^{1}_{m} \in \mathcal{C}_{s}^1$ and $\mathbf{c}^{2}_{m} \in \mathcal{C}_{s}^2$ are the coordinates of the $m$-th pixel in the two sets, and ${\rm Sign}(\cdot)$ is the function that returns the sign (\textit{i.e.}, -1 or +1) of the given value.


 The overall training objective is composed of two primary components. First, there's the standard cross-entropy loss $\mathcal{L}^{ce}_{s}$, which supervises the predicted probability volume $\mathbf{P}_s$ against the ground truth volume for each scale $s$. Second, we introduce the relative consistency loss $\mathcal{L}_{s}^{rc}$, a novel component exclusively applied at the final scale. 
 Formally, the overall training objective of MonoMVSNet is:

\begin{equation}
    \mathcal{L}_{overall} = \sum_{s=0}^{3} \mathcal{L}^{ce}_{s} + \gamma \mathcal{L}^{rc}_{3}.
\end{equation}

\section{Experiments}

\subsection{Datasets and Metrics}

\noindent{\textbf{Dataset.}} DTU \cite{dtu} is a object-centered indoor dataset that includes 128 scenes, with each scene containing 49 or 64 images captured under 7 different lighting conditions. Following the standard practice \cite{mvsnet}, the dataset is divided into training, test, and validation sets. Tanks-and-Temples \cite{tanks} dataset is a large-scale outdoor real-world scene dataset with complex transformations and lighting variations, divided into an intermediate subset with 8 scenes and an advanced subset with 6 scenes. BlendedMVS \cite{blendedmvs} dataset is a large-scale synthetic dataset that includes both indoor and outdoor scenes, offering a training set with 106 scenes and a validation set with 7 scenes.

\noindent{\textbf{Metrics.}} To validate the effectiveness of the proposed MonoMVSNet, we compute metrics for both point clouds and depth. For the point cloud metric, we use the official MATLAB code and evaluation dataset provided by DTU \cite{dtu} to measure the distance between the generated point clouds and the ground truth, reporting accuracy (Acc.), completeness (Comp.), and their average, Overall. Additionally, for the Tanks-and-Temples \cite{tanks} dataset, we evaluate the generated point clouds by uploading them to the official website, reporting F-score in percentage. For the depth metric, we report the Mean Absolute Error (MAE) and depth error ratios at 2mm ($e_2$), 4mm ($e_4$), and 8mm ($e_8$) on DTU.

\begin{table}[!t]
  \centering
  \resizebox{\linewidth}{!}{
    \begin{tabular}{lcccccc}
    \toprule
    Methods & Years & \textbf{Overall}$\downarrow$ & Acc.$\downarrow$ & Comp.$\downarrow$ \\
    \midrule
    Gipuma \cite{gipuma} & ICCV'15 & 0.578 & \textbf{0.283} & 0.873 \\
    COLMAP \cite{colmap} & CVPR'16 & 0.532 & 0.400 & 0.664 \\
    MVSNet \cite{mvsnet} & ECCV'18 & 0.462 & 0.396 & 0.527 \\
    CasMVSNet \cite{casmvsnet} & CVPR'20 & 0.355 & 0.325 & 0.385 \\
    UniMVSNet \cite{unimvsnet} & CVPR'22 & 0.315 & 0.352 & 0.278 \\
    TransMVSNet \cite{transmvsnet} & CVPR'22 & 0.305 & 0.321 & 0.289 \\
    MVSTER* \cite{mvster} & ECCV'22 & 0.313 & 0.350 & 0.276 \\
    WT-MVSNet \cite{wtmvsnet} & NIPS'22 & 0.295 & 0.309 & 0.281 \\
    RA-MVSNet \cite{ramvsnet} & CVPR'23 & 0.297 & 0.326 & 0.268 \\
    GeoMVSNet \cite{geomvsnet} & CVPR'23 & 0.295 & 0.331 & 0.259 \\
    DMVSNet \cite{dmvsnet} & ICCV'23 & 0.305 & 0.338 & 0.272 \\
    ET-MVSNet \cite{etmvsnet} & ICCV'23 & 0.291 & 0.329 & 0.253 \\
    MVSFormer* \cite{mvsformer} & TMLR'23 & 0.289 & 0.327 & 0.251 \\
    DS-PMNet \cite{dspmnet} & AAAI'24 & 0.290 & 0.323 & 0.257 \\
    MVSFormer++* \cite{mvsformer++} & ICLR'24 & 0.281 & 0.309 & 0.252 \\
    GoMVS \cite{gomvs} & CVPR'24 & 0.287 & 0.347 & \textbf{0.227} \\
    \midrule
    \textbf{MonoMVSNet} (Ours) & $N$=5 & \underline{0.278} & 0.313 & \underline{0.243} \\
    \textbf{MonoMVSNet} (Ours) & $N$=9 & \textbf{0.275} & \underline{0.302} & 0.248\\
    \bottomrule
    \end{tabular}%
    }
    \caption{Quantitative results of reconstructed point clouds on the DTU \cite{dtu} evaluation set with distance metrics [$mm$]. The best and second best values are highlighted with \textbf{bold} and \underline{underline}.
    Methods with * denotes using high-resolution images for training.}
  \label{tab:dtu}%
  \vspace{-2mm}
\end{table}%

\begin{table*}[!t]
    \centering
    \resizebox{\textwidth}{!}{
    \begin{tabular}{l|c|c|cccccccc|c|cccccc}
    \hline
    \multirow{2}{*}{Methods} & \multirow{2}{*}{Years} & \multicolumn{9}{c|}{Intermediate subset$\uparrow$} & \multicolumn{7}{c}{Advanced subset$\uparrow$} \\
    \cline{3-18} 
    & & \multicolumn{1}{c|}{\textbf{Mean}} & \multicolumn{1}{c}{Fam.} & \multicolumn{1}{c}{Fran.} & \multicolumn{1}{c}{Hor.} & \multicolumn{1}{c}{Lig.} & \multicolumn{1}{c}{M60} & \multicolumn{1}{c}{Pan.} & \multicolumn{1}{c}{Pla.} & \multicolumn{1}{c|}{Tra.} & \multicolumn{1}{c|}{\textbf{Mean}} & \multicolumn{1}{c}{Aud.} & \multicolumn{1}{c}{Bal.} & \multicolumn{1}{c}{Cou.} & \multicolumn{1}{c}{Mus.} & \multicolumn{1}{c}{Pal.} & \multicolumn{1}{c}{Tem.} \\
    \hline
    COLMAP \cite{colmap} & CVPR'16 & 42.14 & 50.41 & 22.25 & 25.63 & 56.43 & 44.83 & 46.97 & 48.53 & 42.04 & 27.24 & 16.02 & 25.23 & 34.70 & 41.51 & 18.05 & 27.94 \\
    CasMVSNet \cite{casmvsnet} & CVPR'20 & 56.84 & 76.37 & 58.45 & 46.26 & 55.81 & 56.11 & 54.06 & 58.18 & 49.51 & 31.12 & 19.81 & 38.46 & 29.10 & 43.87 & 27.36 & 28.11 \\
    TransMVSNet \cite{transmvsnet} & CVPR'22 & 63.52 & 80.92 & 65.83 & 56.94 & 62.54 & 63.06 & 60.00 & 60.20 & 58.67 & 37.00 & 24.84 & 44.59 & 34.77 & 46.49 & 34.69 & 36.62 \\
    UniMVSNet \cite{unimvsnet} & CVPR'22 & 64.36 & 81.20 & 66.43 & 53.11 & 63.46 & 66.09 & 64.84 & 62.23 & 57.53 & 38.96 & 28.33 & 44.36 & 39.74 & 52.89 & 33.80 & 34.63 \\
    MVSTER \cite{mvster} & ECCV'22 & 60.92 & 80.21 & 63.51 & 52.30 & 61.38 & 61.47 & 58.16 & 58.98 & 51.38 & 37.53 & 26.68 & 42.14 & 35.65 & 49.37 & 32.16 & 39.19 \\
    WT-MVSNet \cite{wtmvsnet} & NIPS'22 & 65.34 & 81.87 & 67.33 & 57.76 & 64.77 & 65.68 & 64.61 & 62.35 & 58.38 & 39.91 & 29.20 & 44.48 & 39.55 & 53.49 & 34.57 & 38.15 \\
    CostFormer \cite{mvsformer} & IJCAI'23 & 64.51 & 81.31 & 65.65 & 55.57 & 63.46 & \underline{66.24} & 65.39 & 61.27 & 57.30 & 39.43 & 29.18 & 45.21 & 39.88 & 53.38 & 34.07 & 34.87 \\
    RA-MVSNet \cite{ramvsnet} & CVPR'23 & 65.72 & 82.44 & 66.61 & 58.40 & 64.78 & \textbf{67.14} & 65.60 & 62.74 & 58.08 & 39.93 & 29.14 & 46.04 & 40.30 & 53.22 & 34.63 & 36.28 \\
    GeoMVSNet \cite{geomvsnet} & CVPR'23 & 65.89 & 81.64 & 67.53 & 55.78 & 68.02 & 65.49 & \underline{67.19} & \underline{63.27} & 58.22 & 41.52 & 30.23 & 46.54 & 39.98 & 53.05 & 35.98 & 43.34 \\
    DMVSNet \cite{dmvsnet} & ICCV'23 & 64.66 & 81.27 & 67.54 & 59.10 & 63.12 & 64.64 & 64.80 & 59.83 & 56.97 & 41.17 & 30.08 & 46.10 & 40.65 & 53.53 & 35.08 & 41.60 \\
    ET-MVSNet \cite{etmvsnet} & ICCV'23 & 65.49 & 81.65 & 68.79 & 59.46 & 65.72 & 64.22 & 64.03 & 61.23 & 58.79 & 40.41 & 28.86 & 45.18 & 38.66 & 51.10 & 35.39 & 43.23 \\
    MVSFormer \cite{mvsformer} & TMLR'23 & 66.37 & 82.06 & 69.34 & 60.49 & 68.61 & 65.67 & 64.08 & 61.23 & 59.53 & 40.87 & 28.22 & 46.75 & 39.30 & 52.88 & 35.16 & 42.95 \\
    DS-PMNet \cite{dspmnet} & AAAI'24 & 64.16 & 81.11 & 63.43 & 60.84 & 62.23 & 64.96 & 61.92 & 61.41 & 57.35 & 39.78 & 28.52 & 44.93 & 39.12 & 51.68 & 33.77 & 40.67 \\
    MVSFormer++ \cite{mvsformer++} & ICLR'24 & \underline{67.18} & \textbf{82.69} & \underline{69.44} & \underline{64.24} & \underline{69.16} & 64.13 & 66.43 & 61.19 & 60.12 & 41.60 & 29.93 & 45.69 & 39.46 & \underline{53.58} & 35.56 & \underline{45.39} \\
    GoMVS \cite{gomvs} & CVPR'24 & 66.44 & \underline{82.68} & 69.23 & \textbf{69.19} & 63.56 & 65.13 & 62.10 & 58.81 & \textbf{60.80} & \underline{43.07} & \textbf{35.52} & \textbf{47.15} & \underline{42.52} & 52.08 & \underline{36.34} & 44.82 \\
    \hline
    \textbf{MonoMVSNet} (Ours) & - & \textbf{68.63} & 82.38 & \textbf{72.89} & 62.80 & \textbf{70.49} & 65.79 & \textbf{68.54} & \textbf{65.54} & \underline{60.59} & \textbf{43.58} & \underline{30.33} & \underline{46.76} & \textbf{42.90} & \textbf{56.31} & \textbf{37.28} & \textbf{47.88} \\
    \hline
    \end{tabular}}
    \caption{Quantitative results on the Tanks-and-Temples benchmark with F-score [\%]. \textbf{Bold} figures represent the best and \underline{underline} figures represent the second best, respectively. The mean refers the average F-score of all scenes.}
    \label{tab:tnt_pc}
    \vspace{-2mm}
\end{table*}

\subsection{Implementation Details}

MonoMVSNet was developed using PyTorch \cite{pytorch} and utilizes the Adam \cite{adam} optimizer. For the DTU \cite{dtu} dataset, the model is trained for 15 epochs using 5-view input images at a resolution of 512$\times$640 with a batch size of 4. The initial learning rate is 0.001, which is halved after the $10$-th, $12$-th, and $14$-th epochs. For the BlendedMVS dataset, we fine-tune the model trained on DTU for 15 epochs using 9-view images at a resolution of 576$\times$768 with a batch size of 4. The initial learning rate is 0.001, which is halved after the $6$-th, $8$-th, $10$-th, and $12$-th epochs. We performe inverse depth sampling \cite{cider} within a depth range from 425$mm$ to 935$mm$, with depth hypotheses in the coarse-to-fine stages as 8-8-4-4, depth intervals of 0.5-0.5-0.5-0.5, and group correlations of 8-8-4-4.

\subsection{Benchmark Performance}

\noindent{\textbf{Evaluation on DTU.}} We compare MonoMVSNet with traditional and learning-based methods. We predict depth using 5-view images with a resolution of 832$\times$1152 and employ a dynamic fusion strategy \cite{d2hcrmvsnet} for point cloud reconstruction. The quantitative results of the point clouds are shown in Tab. \ref{tab:dtu}, where MonoMVSNet achieves the highest overall performance among all methods and ranks second in accuracy, only behind the traditional method Gipuma \cite{gipuma}, which performs poorly overall performance. Notably, MonoMVSNet's memory consumption during inference is 2.01GB, which is significantly better than recent state-of-the-art methods \cite{mvsformer, mvsformer++, gomvs}, as illustrated in Fig. \ref{fig:compare} row 1 (a). The qualitative comparison of depth maps is presented in Fig. \ref{fig:depth_compare_big}, where our method predicts more accurate depth maps in the most challenging scenes, scan48 and scan77.

\noindent{\textbf{Evaluation on Tanks-and-Temples.}} We further evaluate the generalization capability of MonoMVSNet on the Tanks-and-Temples \cite{tanks} benchmark. Consistent with \cite{mvsformer, mvsformer++}, we set the number of input images $N$ = 21 with 2k resolution. A dynamic fusion strategy is adopted for point cloud reconstruction, and the quantitative comparison results are presented in Tab. \ref{tab:tnt_pc}. Our method achieves the highest F-score on both the intermediate and advanced subsets, demonstrating its strong generalization ability.

\begin{table*}[!t]
  \centering
  \small
  \begin{tabular}{lcccccccccccccc}
  \toprule
  \multicolumn{1}{c}{\multirow{2}[0]{*}{Models}} & \multicolumn{3}{c}{Monocular Feature} & \multicolumn{3}{c}{Monocular Depth} & \multirow{2}[0]{*}{Overall$\downarrow$} & \multirow{2}[0]{*}{Acc.$\downarrow$} & \multirow{2}[0]{*}{Comp.$\downarrow$} & \multirow{2}[0]{*}{MAE$\downarrow$} & \multirow{2}[0]{*}{$e_2$$\downarrow$} & \multirow{2}[0]{*}{$e_4$$\downarrow$} & \multirow{2}[0]{*}{$e_8$$\downarrow$}\\
  \cmidrule(lr){2-4} \cmidrule(lr){5-7}
          & \multicolumn{1}{c}{RMF} & \multicolumn{1}{c}{\qiankun{CA}} & \multicolumn{1}{c}{CVPE} & \multicolumn{1}{c}{MDDS} & \multicolumn{1}{c}{EM} & \multicolumn{1}{c}{RCL} & & & & &  \\
  \midrule
   base & & & & & & & 0.303 & 0.323 & 0.282 & 6.53  & 20.49 & 12.85 & 8.61 \\
   \midrule
  (a) & \checkmark & & & & & & 0.293 & 0.322 & 0.263 & 5.86  & 19.08 & 11.61 & 7.55 \\
  (b) & \checkmark & \checkmark & & & & & 0.293 & 0.324 & 0.262 & 5.82  & 19.41 & 11.76 & 7.61 \\
  (c) & \checkmark & \checkmark & \checkmark & & & & 0.288 & 0.317 & 0.259 & 5.72  & 18.69 & 11.33 & 7.37 \\
  \midrule
  (d) & \checkmark & \checkmark & \checkmark & \checkmark & & & 0.292 & 0.318 & 0.265 & 5.44  & 17.55 & 10.60  & 6.82 \\
  (e) & \checkmark & \checkmark & \checkmark & \checkmark & \checkmark & & 0.283 & 0.305 & 0.260 & 5.04  & 16.60  & 9.81  & 6.31 \\
  full & \checkmark & \checkmark & \checkmark & \checkmark & \checkmark & \checkmark & 0.281 & 0.314 & 0.248 & 4.99  & 16.40  & 9.75  & 6.27 \\
  \bottomrule
  \end{tabular}%
  \caption{Ablation study on DTU evaluation set, using normal fusion strategy for point cloud reconstruction. We analyze the effects of the reference monocular feature (RMF), cross attention (CA), cross-view positional encoding (CVPE), monocular guided depth sampling (MDDS), edge map (EM), and relative consistency loss (RCL). }
  \label{tab:ablation}%
  \vspace{-2mm}
\end{table*}%

\begin{table}[!t]
    \centering
    \resizebox{\linewidth}{!}{
    \begin{tabular}{lccccc}
    \toprule
    Feature Encoder & \multicolumn{1}{c}{Overall$\downarrow$} & \multicolumn{1}{c}{MAE$\downarrow$} & Memory$\downarrow$ & \multicolumn{1}{c}
    {Time$\downarrow$} \\
    \midrule
    All (w/o CVPE) & 0.297 & 6.13 & 5.72GB & 0.55s \\
    All (w/ CVPE) & 0.296 & 5.82 & 5.72GB & 0.56s  \\
    Ref. (w/o CVPE) & 0.284 & 5.37 & 2.01GB & 0.24s \\
    Ref. (w/ CVPE) & 0.278 & 4.99 & 2.01GB & 0.25s \\
    \bottomrule
    \end{tabular}
    }
    \caption{Ablation study on feature extraction design.}
    \label{tab:encoder}
    \vspace{-2mm}
\end{table}

\subsection{Ablation Study}
\label{sec:ablation_study}
Ablation study is conducted on DTU \cite{dtu} to verify the effectiveness of each components. Unless otherwise specified, we use 5-view input images with a resolution of 832$\times$1152 and the dynamic fusion strategy \cite{d2hcrmvsnet} for point cloud reconstruction, with all other hyperparameters kept consistent. The overall results of the ablation are presented in Tab. \ref{tab:ablation}. Our method achieves significant improvements in both point cloud and depth metrics, demonstrating the effectiveness of our proposed method.

\noindent{\textbf{Effectiveness of Monocular Feature.}} Compared to the baseline model, introducing monocular features improves the overall point cloud metric from 0.303 to 0.288, and increased the depth metric MAE by 12.4\%. As shown in Tab. \ref{tab:ablation} model (a)(b)(c), using reference monocular features (RMF) improves point cloud completeness from 0.282 to 0.263, indicating that monocular features effectively address areas where matching fails. Additionally, existing cross attention (CA) slightly improves the depth metrics, primarily because traditional positional encoding is not specifically designed for MVS tasks and lacks an understanding of three-dimensional spatial context. However, the introduction of CVPE significantly enhances the performance in both point cloud and depth metrics.

\noindent{\textbf{Effectiveness of Monocular Depth.}} 
The exploitation of monocular depth boosts the overall performance score from 0.292 to 0.281. As shown in Tab. \ref{tab:ablation}, monocular dynamic depth sampling (MDDS) only yields improvements in depth metrics. However, with the introduction of the edge map (EM) shows improvements in both point cloud and depth performance, indicating that excessive monocular sampling could introduce erroneous depth candidates. The relative consistency loss (RCL) balances optimization by enforcing consistency between multi-view depth and monocular depth, further enhancing model performance.

\noindent{\textbf{Feature Extraction Design.}} 
As shown in Tab. \ref{tab:encoder}, we investigate the performance of different feature extraction strategies. ``All''
 refers to extracting monocular features from all input images, while ``Ref.'' indicates extracting monocular features only from the reference image. As illustrated in rows 1 and 3, ``Ref'' outperforms ``All'' in both point cloud and depth metrics, and is more efficient (GPU memory reduced by 64.9\%, run time reduced by 56.4\%). We speculate this is due to the performance gap between different models—excessive monocular features might negatively affect performance. Additionally, the proposed CVPE design further boosts the performance of the ``Ref.'' model with negligible computational cost (rows 3 and 4).

\noindent{\textbf{ViT Variants.}} We conduct the ablation study on different ViT variants of monocular foundation models, as shown in Tab. \ref{tab:vit}. The ViT-S and ViT-B variants exhibit similar performance, while the ViT-L variant shows a decrease in performance. This indicates that the small (ViT-S) variant of monocular foundation models can provide sufficient prior information without relying on highly parameterized base (ViT-B) or large (ViT-L) models.

\noindent{\textbf{Model Efficiency.}} We compare MonoMVSNet with several state-of-the-art methods regarding runtime, GPU memory consumption, and parameter count. The results are summarized in Table \ref{tab:efficiency}. MonoMVSNet consumes the least GPU memory compared to methods without pre-trained models (GeoMVSNet \cite{geomvsnet}, ET-MVSNet \cite{etmvsnet}, and GoMVS \cite{gomvs}) and methods based on pre-trained models (MVSFormer \cite{mvsformer}, MVSFormer++ \cite{mvsformer++}). This advantage stems from its uniquely designed feature extraction strategy. Additionally, MonoMVSNet exhibits significantly faster runtime than GoMVS and possesses substantially fewer trainable parameters than similar models (MVSFormer and MVSFormer++). These results clearly highlight the superior efficiency of our proposed MonoMVSNet.

\begin{table}[!t]
    \centering
    \small
    \begin{tabular}{lccccc}
    \toprule
    Backbones & \multicolumn{1}{c}{Overall$\downarrow$} & Acc.$\downarrow$ & \multicolumn{1}{c}
    {Comp.$\downarrow$} & \multicolumn{1}{c}{MAE$\downarrow$} \\
    \midrule
    ViT-S & 0.278 & 0.313 & 0.243 & 4.99 \\
    ViT-B & 0.278 & 0.294 & 0.262 & 4.98 \\
    ViT-L & 0.281 & 0.311 & 0.251 & 4.94 \\
    \bottomrule
    \end{tabular}
    \caption{Ablation study on different ViT variants.}
    \label{tab:vit}
\end{table}

\begin{table}[!t]
    \centering
    \resizebox{\linewidth}{!}{
    \begin{tabular}{lccccc}
    \toprule
    Methods & Memory & Time & Params(all) & Params(train) \\
    \midrule
    GeoMVSNet \cite{geomvsnet} & 5.21GB & 0.19s & 15.31M & 15.31M \\
    ET-MVSNet \cite{etmvsnet} & 2.91GB & 0.16s & 1.09M & 1.09M \\
    GoMVS \cite{gomvs} & 12.61GB & 0.64s & 1.50M & 1.50M \\
    \midrule
    MVSFormer \cite{mvsformer} & 3.66GB & 0.24s & 28.01M & 28.01M \\
    MVSFormer++ \cite{mvsformer++} & 4.71GB & 0.23s & 126.95M & 39.48M \\
    \midrule
    MonoMVSNet & 2.01GB & 0.25s & 27.68M & 2.89M \\
    \bottomrule
    \end{tabular}
    }
    \caption{Memory, time and parameters comparison per image at a resolution of 832$\times$1152 with 5-view images.}
    \label{tab:efficiency}
    \vspace{-2mm}
\end{table}
\section{Conclusion} In this paper, we present a monocular-priors-guided multi-view stereo network, MonoMVSNet. The monocular feature and monocular depth from the pre-trained monocular foundation model are efficiently and elegantly integrated into feature extraction and depth sampling procedures. In addition, a relative consistency loss is designed to supervise the prediction depth with the monocular depth. With the help of the priors in pre-trained monocular model, MonoMVSNet can predict more accurate depth, especially for the depth-discontinuous regions (\textit{e.g.}, edge regions).  Experimental results demonstrate that our method outperforms state-of-the-art methods on different datasets with less GPU and inference time consumption.

\noindent\textbf{Acknowledgments.} This work was supported by the Beijing Natural Science Foundation (No. L257003), National Natural Science Foundation of China (No. 62402042 and 62227801).


{
    \small
    \bibliographystyle{ieeenat_fullname}
    \bibliography{main}
}

\appendix
\clearpage
\maketitlesupplementary

\appendix

\section{Details of Camera Embedding}
\label{sec:cvpe}
The details of camera embedding module are as follows. The warped feature ($B\times C \times D \times H \times W$) is reshaped to $B\times CD \times H \times W$ and further processed by Conv$\rightarrow$BN$\rightarrow$ReLU layers to produce the output with shape $B\times C \times H \times W$. The camera parameters ($B\times 4 \times 4$) is reshaped to $B \times 16$, processed by a BN layer, and mapped to $B\times C$ by an MLP, which is further reshaped to $B\times C \times 1 \times 1$.  The two features ($B\times C \times H \times W$ and $B\times C \times 1 \times 1$) are added together with brodcast, wich is further processed by  Squeeze-and-Excitation~\cite{selayer}$\rightarrow$Conv layers. The output is $B\times C\times H\times W$, which then add to the corresponding feature.

\section{More Ablation Study}
\label{sec:ablation}

\noindent{\textbf{Compatibility with Monocular Foundation Models.}} To fairly verify the applicability of our method to different monocular foundation models, we replace Depth Anything V2 \cite{dav2} with other ViT-small version monocular foundation models: DINO V2 \cite{dinov2}, Depth Anything V1 \cite{dav1}, and Depth Pro \cite{depthpro}. As shown in Tab. \ref{tab:mfm}, all alternative monocular foundation models exhibit significant improvements in both point cloud and depth performance, among which Depth Anything V2 achieves the best results. This further confirms the generalization capability of our method.

\begin{table}[htbp]
    \centering
    \resizebox{\linewidth}{!}{
    \begin{tabular}{lccccc}
    \toprule
    Models & \multicolumn{1}{c}{Overall$\downarrow$} & Acc.$\downarrow$ & \multicolumn{1}{c}
    {Comp.$\downarrow$} & \multicolumn{1}{c}{MAE$\downarrow$} \\
    \midrule
    Depth Pro \cite{depthpro} & 0.286 & 0.316 & 0.256 & 5.78 \\
    DINO V2 \cite{dinov2} & 0.284 & 0.311 & 0.257 & 5.50 \\
    Depth Anything V1 \cite{dav1} & 0.282 & 0.299 & 0.265 & 5.03 \\
    Depth Anything V2 \cite{dav2} & 0.278 & 0.313 & 0.243 & 4.99 \\
    \bottomrule
    \end{tabular}
    }
    \caption{Ablation study on different monocular foundation models.}
    \label{tab:mfm}
\end{table}

\noindent{\textbf{Number of Views.}} As shown in Tab. \ref{tab:views}, we show the impact of the number of input views. Multi-view information helps alleviate problems such as occlusions, and the reconstruction quality progressively improves with an increasing number of views, saturating at 9 views.

\begin{table}[htbp]
    \centering
    \small
    \begin{tabular}{lccccc}
    \toprule
    $N$ & \multicolumn{1}{c}{Overall$\downarrow$} & Acc.$\downarrow$ & \multicolumn{1}{c}
    {Comp.$\downarrow$} \\
    \midrule
    4 & 0.2825 & 0.315 & 0.250 \\
    5 & 0.2780 & 0.313 & \textbf{0.243} \\
    6 & 0.2765 & 0.309 & 0.244 \\
    7 & 0.2760 & 0.307 & 0.245 \\
    8 & 0.2755 & 0.304 & 0.247 \\
    9 & \textbf{0.2750} & 0.302 & 0.248 \\
    10 & 0.2755 & \textbf{0.299} & 0.252 \\
    \bottomrule
    \end{tabular}
    \caption{Ablation study on the number of input views $N$.}
    \label{tab:views}
\end{table}

\noindent{\textbf{Positional Encoding Design.} We replace Cross-View Positional Encoding (CVPE) with an MLP  to map the traditional 2D positional encoding. As shown in Fig.~\ref{tab:pe}, the Overall$\downarrow$ metric (0.278$\rightarrow$0.285), demonstrating the effectiveness our proposed CVPE.

\begin{table}[htbp]
    \centering
    \begin{tabular}{lccccc}
    \toprule
    Position Encoding & \multicolumn{1}{c}{Overall$\downarrow$} & \multicolumn{1}{c}{MAE$\downarrow$} \\
    \midrule
    w/ 2D PE+MLP & 0.285 & 5.29 \\
    w/ 2D PE+CVPE & 0.278 & 4.99 \\
    \bottomrule
    \end{tabular}
    \caption{Ablation on position encoding design.}
    \label{tab:pe}
\end{table}

\section{More Visualization Results}
\label{sec:visual}

\begin{figure*}[htbp]
  \centering
   \includegraphics[width=1.0\linewidth]{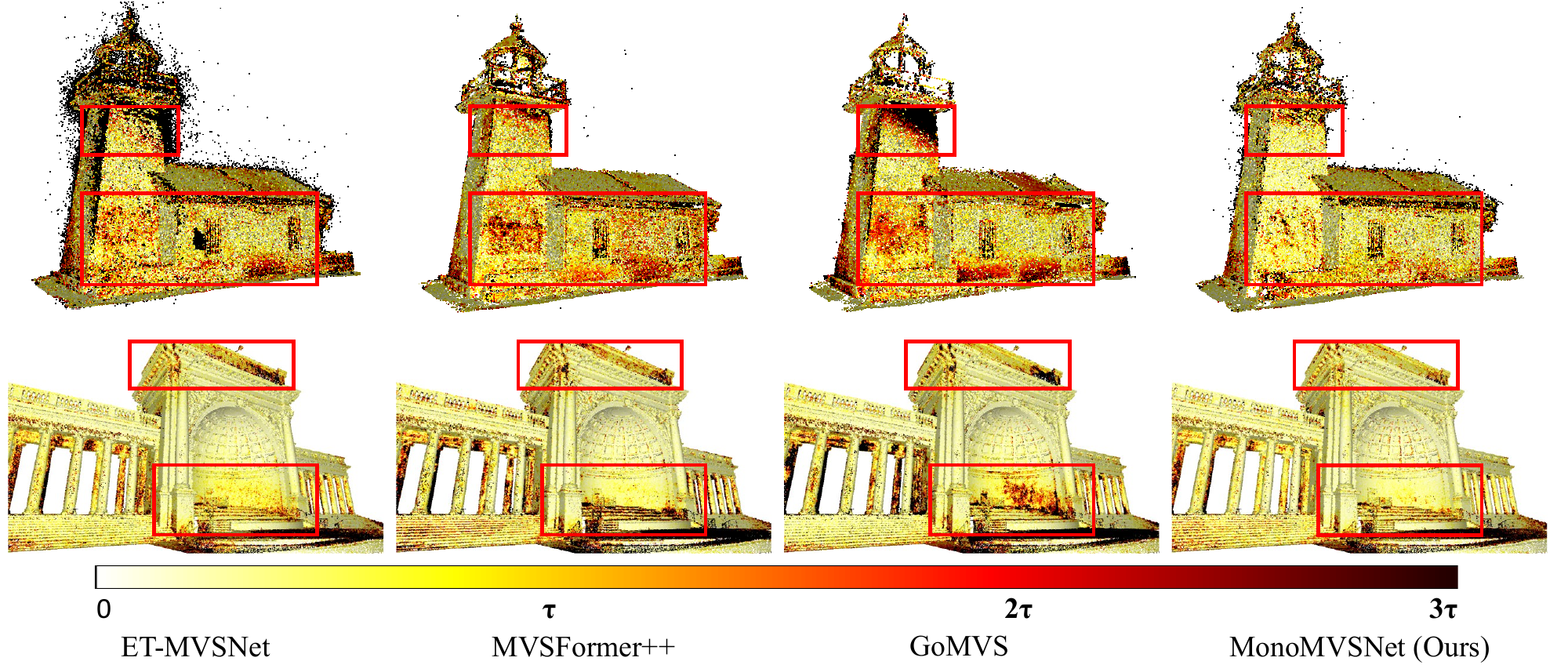}

   \caption{Qualitative comparison of reconstructed point clounds with ET-MVSNet \cite{etmvsnet}, MVSFomrer++ \cite{mvsformer++}, and GoMVS \cite{gomvs} on Tanks-and-Temples \cite{tanks} benchmark. Brighter areas in the figure indicate smaller errors associated with the distance threshold ($\tau$). The top row shows the Precision for the Lighthouse in the advanced subset ($\tau=5mm$), the bottom row shows the Recall for the Temple in the intermediate subset ($\tau=15mm$).}
   \label{fig:tnt_compare}
\end{figure*}

Fig. \ref{fig:tnt_compare} present our method achieves better accuracy and recall in textureless regions and depth discontinuous edge regions on the Tanks-and-Temples benchmark \cite{tanks}. Figure \ref{fig:dtu_depth} presents the depth map comparison results from the ablation experiments on the DTU \cite{dtu} dataset. Figures \ref{fig:dtu_ply} and \ref{fig:tnt_ply} visualize the reconstructed point clouds on the DTU and Tanks-and-Temples benchmark, respectively.

\begin{figure*}[htbp]
  \centering
   \includegraphics[width=1.0\textwidth]{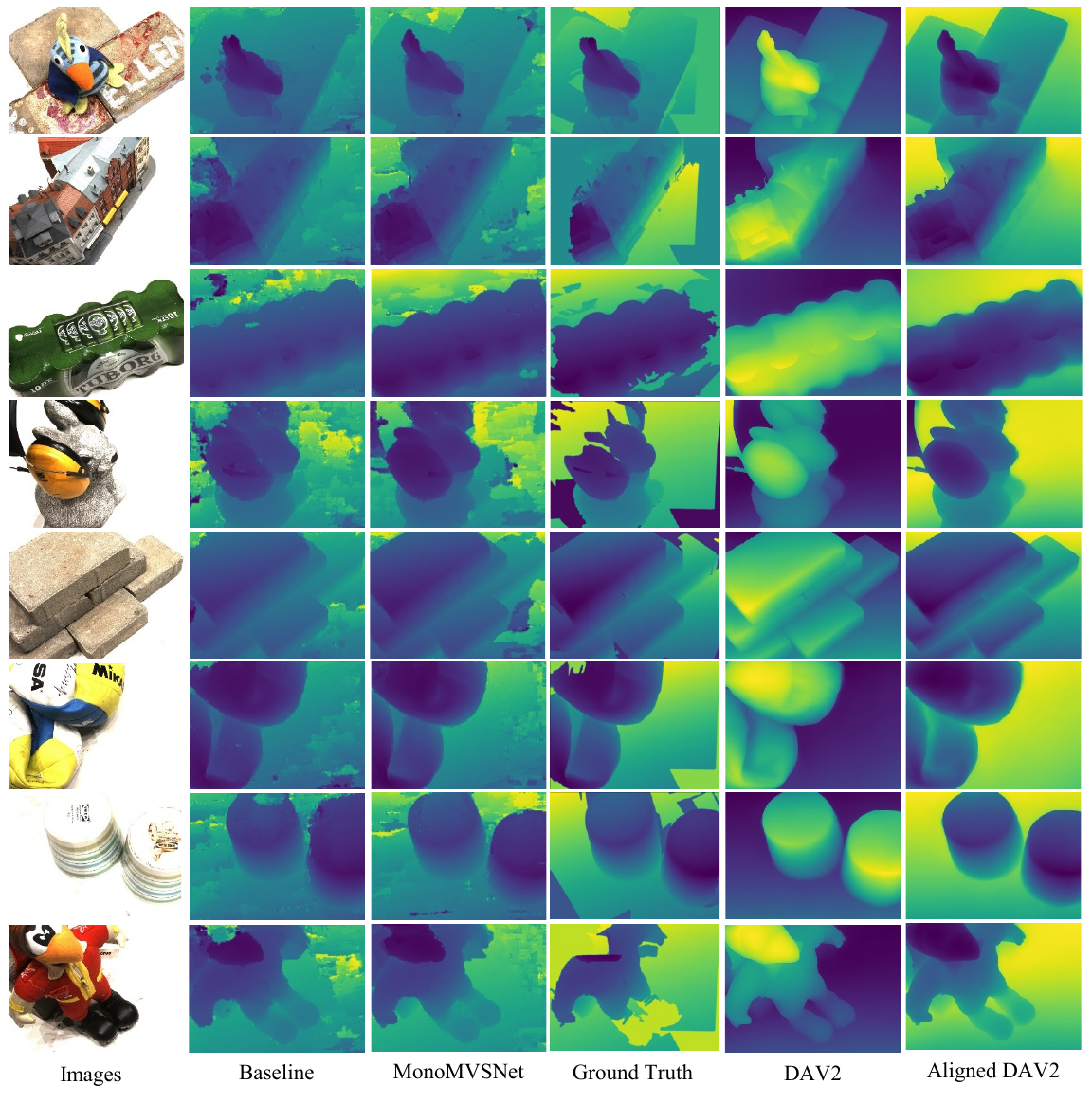}
   \caption{Additional depth maps visualization comparing the Baseline, MonoMVSNet, Ground Truth, Depth Anything V2 (DAV2), and Aligned DAV2 on the DTU \cite{dtu} dataset.}
   \label{fig:dtu_depth}
\end{figure*}

\begin{figure*}[htbp]
  \centering
   \includegraphics[width=1.0\textwidth]{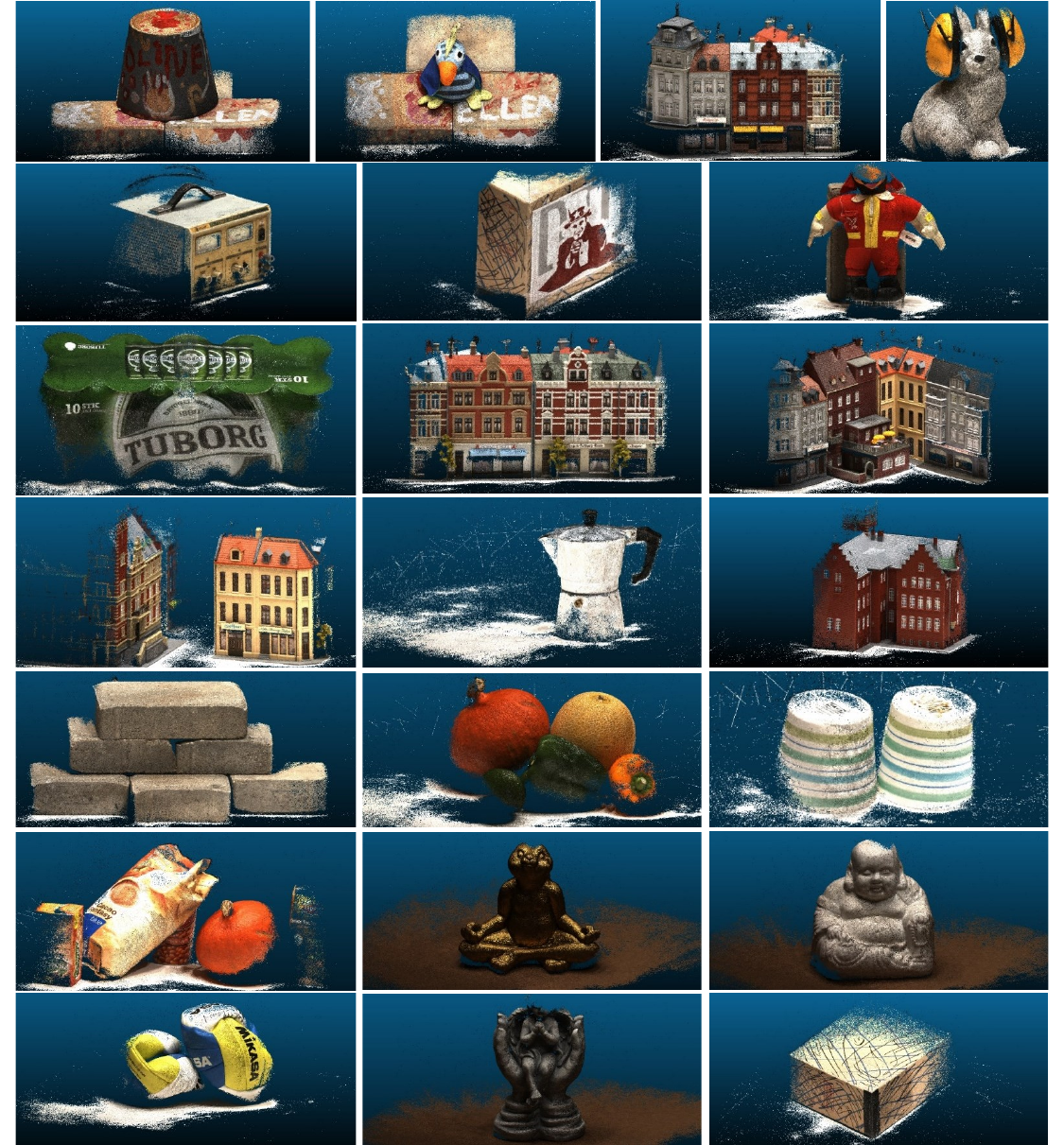}
   \caption{More visualization results of all reconstructed point clouds on the DTU \cite{dtu} dataset.}
   \label{fig:dtu_ply}
\end{figure*}

\begin{figure*}[htbp]
  \centering
   \includegraphics[width=1.0\textwidth]{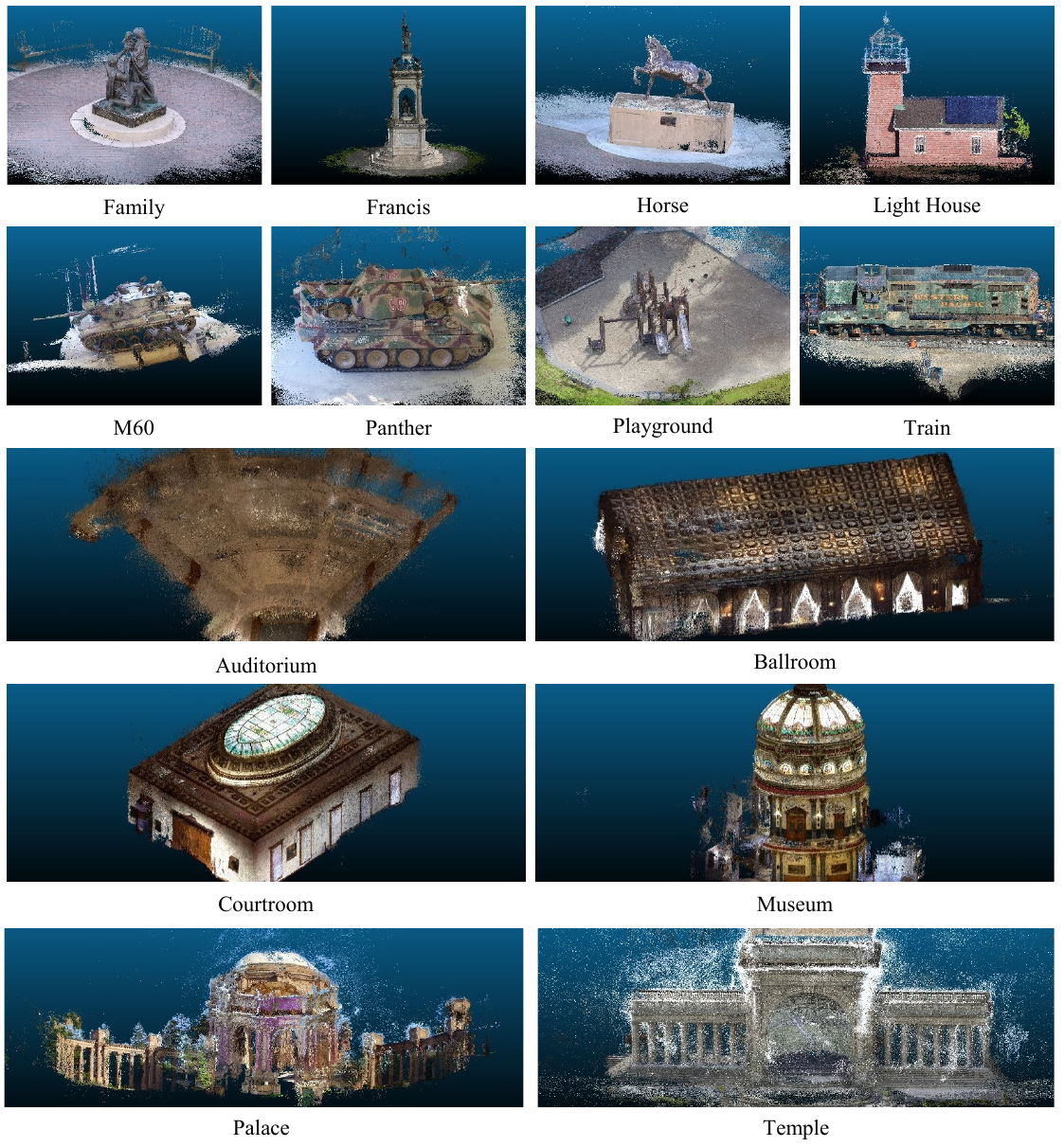}
   \caption{More visualization results of all reconstructed point clouds on the Tanks-and-Temples \cite{tanks} benchmark.}
   \label{fig:tnt_ply}
\end{figure*}

\end{document}